# Prediction of Breast Cancer Recurrence Risk Using a Multi-Model Approach Integrating Whole Slide Imaging and Clinicopathologic Features

Manu Goyal*, PhD, Department of Biomedical Data Science, Dartmouth College, Hanover, NH, USA

Jonathan D. Marotti, MD, Department of Pathology and Laboratory Medicine, Dartmouth Health, Lebanon, NH, USA

Adrienne A. Workman, MD, Department of Pathology and Laboratory Medicine, Dartmouth Health, Lebanon, NH, USA

Elaine P. Kuhn, MD, Geisel School of Medicine, Dartmouth College, Hanover, NH, USA

Graham M. Tooker, MD, Department of Radiology, Dartmouth Health, Lebanon, NH, USA

Seth K. Ramin, MPH, Geisel School of Medicine, Dartmouth College, Hanover, NH, USA

Mary D. Chamberlin, MD, Department of Medicine, Dartmouth Health, Lebanon, NH, USA

Roberta M. diFlorio-Alexander, MD, MS, Department of Radiology, Dartmouth Health, Lebanon, NH, USA

Saeed Hassanpour, PhD, Departments of Biomedical Data Science, Computer Science, and Epidemiology, Dartmouth College, Hanover, NH, USA

**Short title:** A Multi-Model Approach for Breast Cancer Recurrence Risk Prediction

**Conflicts of interest:** The authors have no financial, professional, or personal conflicts of interest.

**Funding sources:** This research was supported in part by grants from the US National Library of Medicine (R01LM012837 & R01LM013833) and the US National Cancer Institute (R01CA249758).

**Number of text pages:** 16

**Number of Figures:** 4

**Number of tables:** 4

*Corresponding Author: Manu Goyal, 1 Medical Center Drive, Williamson Translational Research Building, Lebanon, NH, 03756. Manu.Goyal@Dartmouth.edu, +1 (603) 678-9293

**Abstract:**

Breast cancer is the most common malignancy affecting women worldwide and is notable for its morphologic and biologic diversity, with varying risks of recurrence following treatment. The Oncotype DX Breast Recurrence Score test is an important predictive and prognostic genomic assay for estrogen receptor-positive breast cancer that guides therapeutic strategies; however, such tests can be expensive, delay care, and are not widely available. The aim of this study was to develop a multi-model approach integrating the analysis of whole slide images and clinicopathologic data to predict their associated breast cancer recurrence risks and categorize these patients into two risk groups according to the predicted score: low and high risk. The proposed novel methodology uses convolutional neural networks for feature extraction and vision transformers for contextual aggregation, complemented by a logistic regression model that analyzes clinicopathologic data for classification into two risk categories. This method was trained and tested on 993 hematoxylin and eosin-stained whole-slide images of breast cancers with corresponding clinicopathological features that had prior Oncotype DX testing. The model's performance was evaluated using an internal test set of 198 patients from Dartmouth Health and an external test set of 418 patients from the University of Chicago. The multi-model approach achieved an AUC of 0.92 (95% CI: 0.88–0.96) on the internal set and an AUC of 0.85 (95% CI: 0.79–0.90) on the external cohort. These results suggest that with further validation, the proposed methodology could provide an alternative to assist clinicians in personalizing treatment for breast cancer patients and potentially improving their outcomes.

**Introduction**

Breast cancer remains a leading cause of cancer-related mortality among women worldwide [1]. The American Cancer Society estimated approximately 284,200 new cases of breast cancer diagnosed in the United States in 2023, resulting in an estimated 44,130 deaths [2]. The prognosis and treatment of breast cancer are influenced by various factors, including age, ethnicity, tumor size, and molecular subtype (i.e., tumor biology) [3]. The Oncotype DX Breast Recurrence Score test has become a crucial tool for clinical decision-making, particularly in hormone receptor-positive, HER2-negative early-stage breast cancer, which represents about 70% of new cases in the United States [4].

The Oncotype DX test is a genomic assay that assesses the expression of 21 genes within a tumor sample to calculate a recurrence score ranging from 0 to 100. The recurrence score categories are typically divided into two groups: 1) Low risk (scores 0-25), which is considered to have a low/moderate risk of cancer recurrence. These patients are less likely to benefit from chemotherapy and may avoid the potential side effects of these treatments. 2) High risk (scores 26-100), which is considered to have a significant risk of cancer recurrence. Chemotherapy is often recommended for these patients as the potential benefits typically outweigh the risks.

The Oncotype DX Breast Recurrence Score helps clinicians predict the likelihood of cancer recurrence and the potential benefit of chemotherapy [5]. By evaluating cancer-related and reference genes, the test provides an individualized risk assessment that guides adjuvant chemotherapy and hormone therapy decisions [6]. Consequently, this test has been incorporated into clinical guidelines and has become a standard practice in many countries [7]. However, such genomic assays can be expensive, potentially delay care, and are not widely available and not always covered by insurance in low-resource settings. The focus of this study was to develop Artificial Intelligence (AI) methods for binary classification of high- and low-risk breast cancer recurrence that can assist clinicians in identifying patients who have a clear indication for chemotherapy (high risk) and those who could potentially avoid it (low risk), thus simplifying and streamlining treatment decisions.

In medical image analysis, AI and deep learning have achieved notable success in diagnosing and prognosticating chronic diseases, particularly in oncology [9-15]. Recent advances in AI and machine learning have shown potential in accurately predicting breast cancer recurrence risks based on Oncotype DX testing [16-20]. In a recent study, Howard et al. integrated machine learning models with clinical data, achieving an Area Under the ROC Curve (AUC) of 0.83 in an external cohort [16]. Studies by Romo-Bucheli et al. demonstrated that automated quantification of tubule nuclei [17] and mitotic activity [18] could differentiate between high and low Oncotype DX risks. However, these studies' accuracy analyses did not account for patients with Oncotype DX scores ranging from 18 to 29, limiting their clinical utility. In another study, quantitative analysis of nuclear histomorphologic features yielded an AUC of 0.65 for identifying cases with high Oncotype DX risk [19]. A proprietary tile-based convolutional neural network model by Cho et al., interpreting cellular, structural, and tissue-based features from image tiles, achieved an AUC of 0.75 for predicting high Oncotype DX risk [20].

It should be noted that these previous AI-based approaches often depended on manual annotations or conventional image analysis methods that do not fully consider the spatial context and relationships among the diverse morphological features within a whole-slide image [23-25]. This lack of utilizing more modern machine learning techniques for Oncotype DX breast recurrence scoring highlights the need for innovative solutions to enhance diagnostic accuracy, efficiency, and personalization in breast cancer treatment. These considerations motivated the current study, which sought to bridge such gaps. The development of novel and accurate AI models that use clinicopathologic and digital imaging data for predicting genomic-based prognostic scores, such as breast cancer recurrence risks based on Oncotype DX test, can have a major positive impact on the accessibility of high-quality cancer care, especially in resource-limited healthcare settings. These models, when deployed as online servers, require only a limited computational resources and connectivity capabilities, and can be readily accessed at a fraction of the cost of the genomic assay tests with no latency.

Our study introduces OncoDHNet, a novel deep-learning model that incorporates recent advancements in masked pre-training techniques for analyzing whole-slide images, eliminating the need for manual annotation on histopathology slides [26, 27]. OncoDHNet takes advantage of transformer architectures to analyze and reconstruct image patches, facilitating an integration of both local and global histological features. In addition to histopathology images, a logistic regression model trained on corresponding clinicopathologic features is utilized to improve predictive accuracy, akin to a study by researchers from the University of Tennessee [28]. This multi-modal approach uses whole slide images and their corresponding clinicopathologic features to categorize breast cancer into two distinct risk categories based on predicted recurrence scores. This methodology combines deep learning-based advanced capabilities with interpretable logistic regression analysis, promising to aid oncologists in creating more accurate and personalized treatment protocols and enhancing patient care.

**Materials and Methods**

**Datasets**

In this study, we used an internal dataset of hematoxylin and eosin (H&E) stained slides and clinicopathologic data from Dartmouth Health to train and test the multi-modal approach, which was then externally validated using a publicly available database known as The University of Chicago (UC) database [16]. The use of human subject data in this project received approval from the Dartmouth Health Institutional Review Board (IRB) with a waiver of informed consent. The details of these datasets are included below.

**Internal Dataset.** This dataset comprises 990 Whole Slide Images (WSIs) and corresponding clinicopathologic data from 981 patients, provided by the Department of Pathology and Laboratory Medicine at Dartmouth Health, a tertiary academic care center in Lebanon, New Hampshire. Since this study focuses on only ER-positive patients, after excluding 6 ER-negative patients, the internal dataset constitutes of 984 slides from 975 patients. The H&E-stained WSIs were obtained from the same tissue blocks that were previously sent for Oncotype DX breast recurrence testing from 2013-2022. The Oncotype DX breast recurrence score for each WSI in the internal dataset was retrieved from the associated reports. The threshold of 26 on Oncotype DX breast recurrence scores was used to establish the low and high recurrence categories for these patients. The slides were digitized using Leica Aperio AT2 and CS2 scanners at 40× magnification (0.25 μm/pixel). Clinicopathologic data was extracted from pathology reports and included: patient age; tumor size; tumor grade; histologic type, estrogen receptor (ER) status; progesterone receptor (PR) status; and human epidermal growth factor receptor 2 (HER2) status [28]. Further details of the clinicopathologic data are shown in Table 1.

**Table 1.** The demographic information of the internal and UC cohorts. In the internal cohort, there was one patient for whom the PR and four patients for whom HER2 status was not available, whereas in the external cohort, the ER status was missing for one patient, and the HER2 status was not available for nine patients. Abbreviations: ER Estrogen Receptor, PR Progesterone Receptor, HER2 Human Epidermal Growth Factor Receptor 2, SD Standard Deviation.

| Clinicopathologic Features | | Internal Cohort | UC Cohort |
|---|---|---|---|
| No. of patients | | 981 | 427 |
| Age, mean (SD) | | 61.6 (11.1) | 56.3 (10.6) |
| Histologic Subtype, n (%) | Invasive Ductal Carcinoma | 757 (77.1) | 302 (70.7) |
| | Invasive Ductal and Lobular Carcinoma | 28 (2.9) | 40 (9.4) |
| | Invasive Lobular Carcinoma | 166 (16.9) | 71 (16.6) |
| | Others | 30 (3.1) | 14 (3.3) |
| Tumor Grade, n (%) | 1 | 252 (25.6) | 68 (15.9) |
| | 2 | 528 (53.8) | 279 (65.3) |
| | 3 | 201 (20.3) | 80 (18.7) |
| Tumor Size (mm), mean (SD) | | 20.4 (15.6) | 21.0 (16.2) |
| ER Status, n (%) | Positive | 975 (99.3) | 417 (97.9) |
| | Negative | 6 (0.7) | 9 (2.1) |
| PR Status, n (%) | Positive | 900 (91.7) | 374 (87.6) |
| | Negative | 80 (8.3) | 53 (12.4) |
| HER2 Status, n (%) | Positive | 23 (2.3) | 15 (3.5) |
| | Negative | 954 (97.2) | 410 (96.5) |
| Oncotype DX Breast Recurrence Score, mean (SD) | | 18.5 (11.7) | 18.6 (10.1) |

**The University of Chicago dataset**. The UC dataset comprises 487 digitized H&E tumor sample slides with corresponding clinicopathologic data, collected from 427 patients [16]. The 9 ER-negative patients were omitted from external cohort. The final external cohort includes 389 slides from 343 patients with low Oncotype DX breast recurrence scores (<26) and 87 slides from 75 patients with high Oncotype DX breast recurrence scores (>=26). This publicly released dataset does not contain the full WSIs; however, it only includes specific patches extracted from each WSI, along with their corresponding locations. Hence, we used a clustering algorithm to organize these patches into distinct regions, which were then utilized for testing our model. The distribution of the two breast cancer recurrence risk groups (low breast cancer recurrence risk and high breast cancer recurrence risk) for both datasets is summarized in Table 2.

**Table 2.** Distribution of low- and high-risk cases in the internal and UC datasets.

| Oncotype DX Breast Recurrence | Internal Dataset | UC Dataset |
|---|---|---|
| Low Breast Cancer Recurrence Risk, n (%) | 821 Slides (83.2%) from 813 patients | 389 (81.6%) Slides from 343 patients |
| High Breast Cancer Recurrence Risk, n (%) | 163 Slides (16.8%) from 162 patients | 87 (18.4%) Slides from 75 patients |
| Total | 984 Slides from 975 patients | 476 Slides from 418 patients |

**Dataset Partitioning for Model Development and Evaluation.** The slides from the internal cohort were divided into subsets for model development and evaluation, as shown in Table 3. A subset of 688 slides of 680 patients from the internal dataset (approximately 70% of the entire dataset) was used as the training set, 98 internal slides from 97 patients (roughly 10% of the dataset) were employed as the development set, and a collection of 198 internal slides from 198 patients (about 20% of the dataset) constituted the internal testing set. To further extend our evaluation, all the slides from the UC database were utilized as an external test set. This external test set was instrumental in assessing the generalizability of our approach. Careful attention was paid to maintaining patient consistency within each set, ensuring that slides from the same patients were contained within the same partition to prevent information leakage. Additionally, the breast cancer recurrence risk category composition was balanced in the training/development/testing split to maintain coherence and comparability across the different partitions as shown in Table 3.

**Table 3.** Distribution of patients across different partitions for model development and evaluation.

| Oncotype DX Breast Recurrence Score | Internal Training Dataset | Internal Development Dataset | Internal Testing Dataset | External (UC) Testing Dataset | Combined |
|---|---|---|---|---|---|
| Low Oncotype DX Breast Recurrence (Score of 0-25) | 568 | 81 | 164 | 343 | 1156 |
| High Oncotype DX Breast Recurrence (Score of >=26) | 112 | 16 | 34 | 75 | 237 |
| Total | 680 | 97 | 198 | 418 | 1393 |

**The Proposed Multi-Model Approach**

In this study, a new multi-model approach was designed to integrate clinicopathologic features derived from pathology reports with morphological features from WSI. We used a logistic regression model to analyze the clinicopathologic features, while developing OncoDHNet, a transformer-based architecture inspired by previous work from our team [26, 27], to analyze the whole-slide images. OncoDHNet employs Convolutional Neural Network (CNN) and transformer techniques to capture both detailed and broader features within WSIs by training the model to accurately reconstruct any obscured or incomplete image areas. This functionality is essential for identifying important histomorphological characteristics to categorize WSIs into low and high breast cancer recurrence risk groups. Furthermore, this technique highlights critical regions within the WSI images that significantly influence the classification outcome. By highlighting these key regions, pathologists can discern which parts of the WSI contribute most to the model's outcome, thereby enhancing the transparency and confidence on the model's findings.

OncoDHNet combines detailed features extracted from image patches of WSIs using a pre-trained CNN model with broader contextual features identified by a vision transformer model. OncoDHNet utilizes the strengths of both CNNs [29] and Vision Transformers [30] to capture both micro-level details and macro-level patterns within the tissue samples. Initially, selected WSI regions are divided into 224×224-pixel patches, which are processed through a CNN with a ResNet-18 architecture to extract feature vectors. These feature vectors can capture histopathological information, such as cellular irregularities and tissue architecture. The feature vectors are then passed to a pre-trained transformer model, developed in our previous work [27], to fill in the artificially masked regions of a whole-slide image, thereby gaining the ability to capture subtle histopathological features. This model assesses 50 randomly selected 4480×4480-pixel regions from a WSI, synthesizing 'class tokens' for each region. The average of these class tokens forms a holistic representation of the slide, which is used to determines the final breast cancer recurrence risk category.

A logistic regression model was also trained on the clinicopathologic features of the internal dataset as listed in Table 1 [28]. The logistic regression was fine-tuned to optimize its corresponding hyperparameters such regularization strengths and penalty types. In this optimization, a grid search method was used, with a 5-fold cross-validation on the development dataset, focused on maximizing the AUC for predicting breast cancer recurrence risk categories.

Finally, the combined model was built to integrate predictions from the OncoDHNet model and logistic regression. This approach searches for optimal threshold settings for both models to maximize the AUC metric and ensure the most effective integration of image based OncoDHNet predictions with clinicopathologic feature analysis. The objective is to identify an optimal threshold for each model's confidence scores. This is achieved through a process that evaluates various potential cut-off points, aiming to find the one that best discriminates between low and high-risk categories, as indicated by the AUC. The entire training process of the proposed methodology, including the integration of OncoDHNet and logistic regression, is depicted in Figure 1.

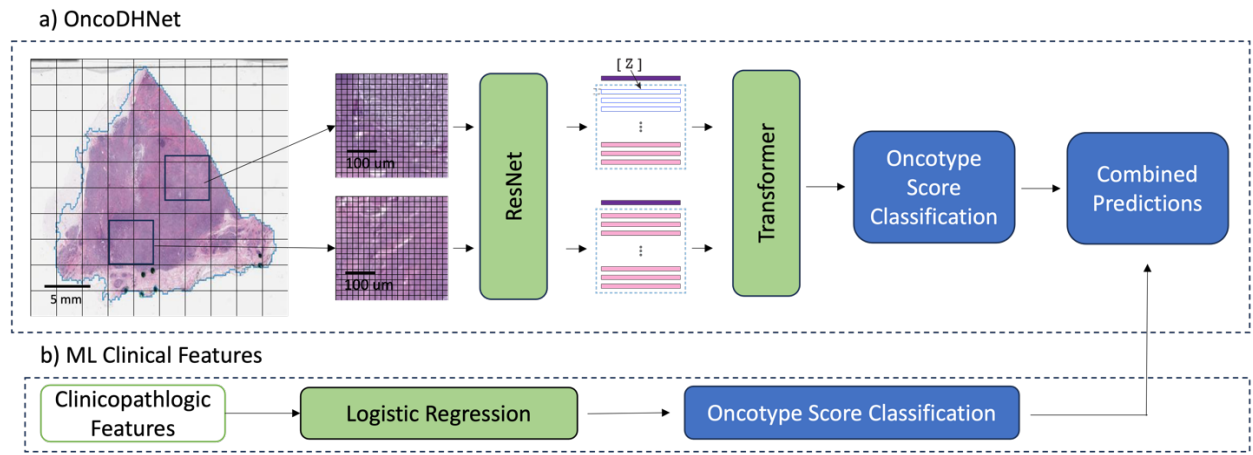

**Figure 1.** The overview of proposed multi-model approach including a) OncoDHNet architecture for WSI and b) logistic regression for clinicopathologic features for predicting breast cancer recurrence risk.

**Evaluation Metrics and Statistical Analysis**

To assess the performance and effectiveness of our methodology, our trained model was evaluated on a held-out test set of 198 patients in our internal dataset and an additional 418 patients in the external UC dataset. The evaluation relied on the AUC, a key metric for classification. The 95% confidence intervals (CIs) were calculated using the bootstrap resampling technique with 10,000 iterations. The AUC score measures the model's ability to distinguish between lower (Oncotype DX score below 26) and higher (Oncotype DX score 26 or above) breast recurrence score groups; a higher AUC value signifies better performance.

**Results**

In our approach, the model was first trained to predict the individual Oncotype DX breast recurrence score for each slide, i.e., a number between 0 and 100, achieving an $R^2$ score of 0.42. This pre-trained model was then used in the second phase for classification between lower and higher risk categories. Table 4 show our detailed evaluation results in this study. The AUC scores and their corresponding 95% confidence intervals for our proposed methods in distinguishing low versus high breast cancer recurrence risks using WSIs for different testing cohorts are presented in this table. The internal test cohort consisted of 198 slides from 198 patients, whereas in the external cohort included 476 slides corresponding to 418 patients. The AUCs were calculated on a per-patient basis by averaging the scores across multiple WSIs, if more than one is available, for each patient. OncoDHNet achieved an AUC score of 0.87 (95% CI: 0.81 to 0.92) on the internal test set. The logistic regression model, trained on clinicopathologic features, also performed notably, with an AUC of 0.84 (95% CI: 0.76–0.90) on the internal test set. The combined model, integrating the OncoDHNet image-based approach with the clinicopathologic insights, showed superior performance with an AUC of 0.92, highlighting the advantage of a multi-modal analytic approach.

In the external UC test set, OncoDHNet and the logistic regression model achieved AUC scores of 0.84 (95% CI: 0.78–0.89) and 0.78 (95% CI: 0.72–0.84), respectively. The combined model again demonstrated superior performance, with an AUC of 0.85 (95% CI: 0.79–0.90). The receiver operating characteristic (ROC) curves for both internal and external datasets are depicted in Figure 2.

**Table 4.** Evaluation of the performance of OncoDHNet, Logistic Regression and Combined Model on internal and external test sets. The 95% CI of each metric is included in parentheses.

| Datasets | Methods | AUC |
|---|---|---|
| Internal Test Set | OncoDHNet | 0.87 (0.81–0.92) |
| | Logistic Regression | 0.84 (0.76–0.90) |
| | Combined Model | 0.92 (0.88–0.96) |
| UC Test Set | OncoDHNet | 0.84 (0.78–0.89) |
| | Logistic Regression | 0.78 (0.72–0.84) |
| | Combined Model | 0.85 (0.79–0.90) |

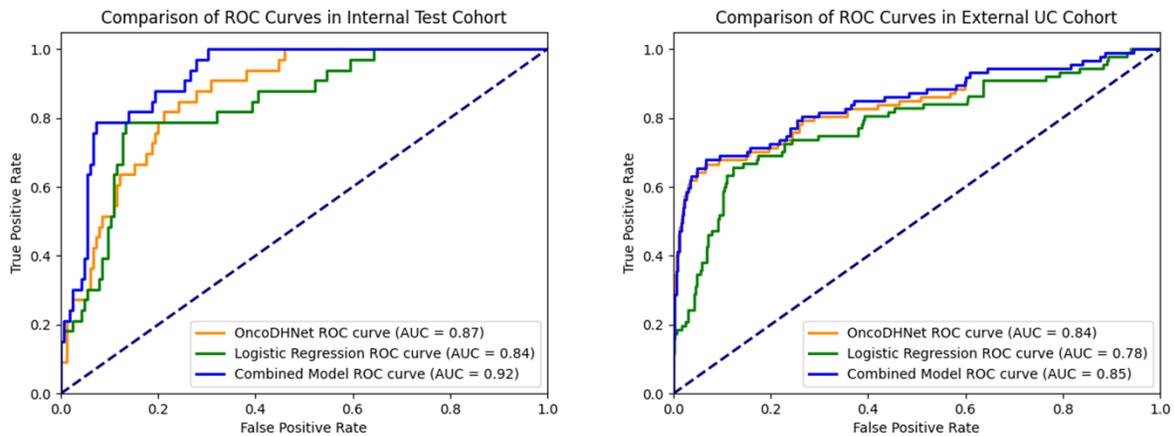

**Figure 2.** ROC curves demonstrate the performance of OncoDHNet based on histopathology images, Logistic Regression on clinicopathologic features, and a Combined Model on both the internal and external (UC) test sets.

For the logistic regression analysis, which focuses on clinicopathologic features to predict breast cancer recurrence risk categories, the most influential features were determined by considering the values of the model coefficients (Figure 3). In this analysis, tumor grade emerged as the most influential factor for predictive analysis. It was followed by PR and HER2 status, denoting the presence of progesterone receptors and Human Epidermal Growth Factor Receptor 2, as the second and third most significant predictors, respectively. Tumor size, patient age, and histologic types also contribute to the model's predictive power, albeit to a lesser degree compared to the first three significant predictors. Notably, all these important features have a positive correlation with a high-risk outcome. ER status, however, is shown to have no importance in the model as this study focuses only on ER positive patients.

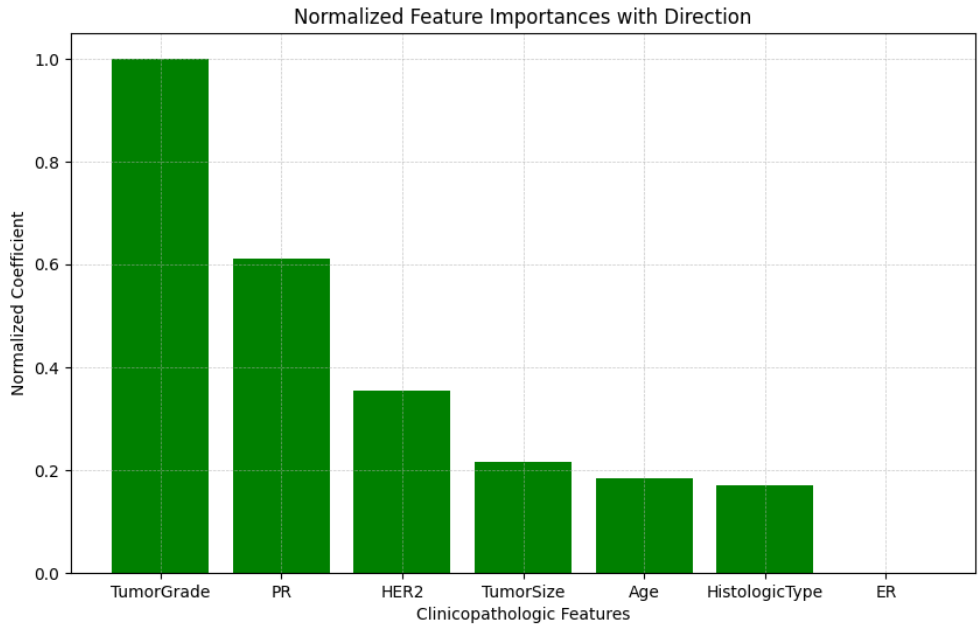

Figure 3. The Feature importance of logistic regression model trained on clinicopathologic features, where a green bar represents influence in positive directions. Abbreviations: ER Estrogen Receptor, PR Progesterone Receptor, HER2 Human Epidermal Growth Factor Receptor 2.

We evaluated the performance of our proposed models (OncoDHNet, logistic regression, and a combined model) on both internal and external test datasets, using the Mann-Whitney U test with Bonferroni correction [31] for multiple comparisons of AUC scores. Our analysis revealed that in the internal test set, OncoDHNet significantly outperformed logistic regression (p-value < 0.01), and the combined model was statistically superior to both OncoDHNet and logistic regression (p-value < 0.01). Similarly, in the external test set, OncoDHNet demonstrated a significant advantage over logistic regression (p-value < 0.01), while the combined model surpassed both OncoDHNet and logistic regression (p-value < 0.01), confirming its robustness and superior discriminative ability across datasets.

**Visualization**

Figure 4 shows randomly selected examples from the internal test dataset, depicting the attention maps generated by the OncoDHNet model. These maps visually highlight the regions that highly influence (red) and minimally influence (blue) OncoDHNet's outcome when predicting the breast cancer recurrence risk category. Of note, since there are no WSI images available for the UC cohort, the attention maps were only generated on the internal test set.

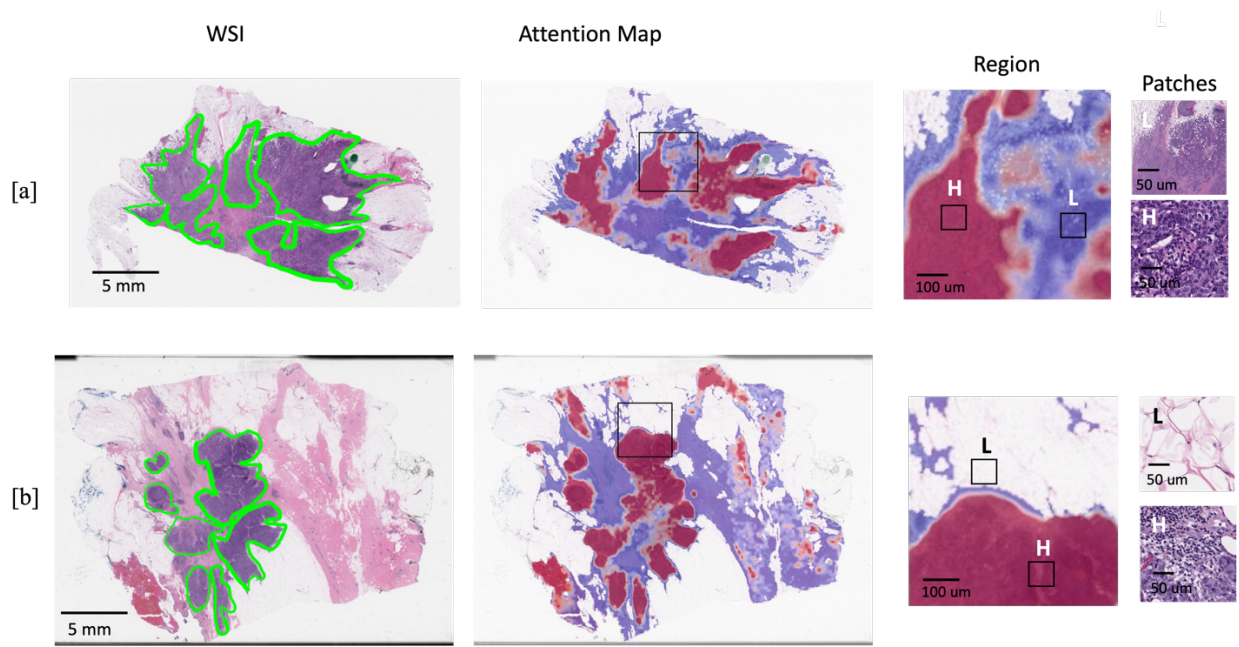

**Figure 4.** Display of attention maps of OncoDHNet for predicting breast cancer recurrence risk categories on randomly selected WSIs from the Internal Test set. The high-attention areas are marked in red (H), while the areas of lower attention are indicated in blue (L). The first column, labeled "WSI," shows the original whole slide image (green pen markings indicates tumor regions and were made by a pathologist during the initial case review and verification).

**Discussion**

This study focuses on a multi-modal approach to integrate clinicopathologic and histopathological features for predicting breast cancer recurrence risk categories to inform breast cancer treatment. It introduces a novel and efficient strategy for assessing the potential benefits of chemotherapy and the risk of recurrence in ER-positive breast cancer, currently evaluated by the Oncotype DX genomic test. The high cost and logistical challenges of this test have spurred the need for more accessible alternatives.

This research integrates a linear model (logistic regression) that analyzes clinicopathologic features with OncoDHNet, a transformer model trained on histopathological features, to predict the breast cancer recurrence risk category for breast cancer. This new approach aims to provide a comprehensive evaluation by harnessing both clinicopathologic data and detailed histopathological image analysis, streamlining the process of determining the breast cancer recurrence risk categories based on Oncotype DX testing, which is crucial for guiding treatment decisions in breast cancer care.

The logistic regression model processes important clinicopathologic features, while OncoDHNet uses advanced image processing techniques to assess WSIs for morphological patterns associated with cancer severity. By combining the strengths of both models, the proposed multi-modal approach can accurately classify breast cancer cases into high or low risk for recurrence as indicated by Oncotype DX breast recurrence scores, potentially reducing the need for costly genomic testing.

This multi-model strategy allows for a more robust and nuanced prediction, leveraging the logistic regression model's clinicopathologic insights and OncoDHNet's visual analysis capabilities. The combination of these two models provides a more detailed prognostic assessment, offering an opportunity to personalize treatment strategies for breast cancer patients more effectively. This tool can streamline the diagnostic and treatment planning processes, potentially improving how clinicians approach breast cancer prognosis.

OncoDHNet is designed to perform an in-depth analysis of the complex morphological patterns present in high-resolution histology slides. Using vision transformers, it captures both subtle and significant features indicative of breast cancer's aggressiveness. This model not only emulates the predictions made by genomic assays but also provides a practical alternative for prognosis in settings where such genomic tests are not feasible due to resource limitations, including the patients lacking access to specialized cancer care.

Overall, this study evaluates the effectiveness of three models for determining the low- and high-breast cancer recurrence risk categories. The logistic regression model, trained on seven clinicopathologic features, is efficient and offers a high degree of interpretability with feature importance discussed later in this section. However, this computational simple approach achieved lower AUC scores in both the internal and the external cohort, suggesting potential limitations in determining the score with only clinicopathologic data leading to a weaker generalizability and predictive accuracy especially in external cohort. On the other hand, OncoDHNet requires significant amounts of data and computational resources to train the model, however, provides a better AUC score in internal and external cohorts which suggests the capability of model to capture the complex pattern in histopathological data to discriminate between recurrence risk categories in both cohorts. Finally, the combined model provides superior performance in both cohorts compared to logistic regression and OncoDHNet alone. This multi-modal strategy, while the most effective across both data sets, requires a careful integration of the two modalities.

It is important to emphasize that no WSIs were publicly available in the external UC dataset and only patches were available and used in the external evaluation. WSIs provide a comprehensive view of the tissue sample, allowing models to analyze the full context of the tissues, which is particularly important in cancer where heterogeneity within a single slide can be significant. The ability to analyze whole slides would likely enable more accurate modeling of spatial relationships and morphological patterns in the external cohort that are critical in cancer diagnosis and prognosis. The use of patches can limit the model to the information contained within those smaller extracted regions. Although these patches were selected from critical areas, they may not encompass all the diagnostic features present in a WSI. For instance, subtle indicators of cancer severity or small regions of aggressive disease could be missed if they fall outside the selected patches. In the absence of WSIs in the external dataset, our methodology was adapted by using a clustering algorithm to group the patches into unified regions, which allowed the model to infer some level of spatial context. Of note, this approach is inherently more noisy and less informative compared to the insights that could be extracted from whole slides. Therefore, while the study achieved promising results with the available external dataset, the full potential of our image analysis methodology, especially in terms of spatial analysis and comprehensive feature extraction, could likely be better realized with the availability of WSIs. In future work, we will plan to obtain new WSI datasets from other institutions to further evaluate and refine the model and improve its predictive accuracy.

For the logistic regression model, feature importance was derived to determine which clinicopathologic features were most influential in predicting breast cancer recurrence risk categories. Tumor grade, being the most influential factor in our model, indicates that the model places significant emphasis on tumor cell aggressiveness. In clinical practice, higher-grade tumors are often more aggressive and may require more intensive treatment. PR status and HER2 status were the second and third most significant features, respectively. Oncotype DX breast recurrence test is typically not performed on HER2-positive tumors. A review of the tumors considered HER2 positive in our cohort found that some were HER2 positive on biopsy, but subsequently deemed to be HER2 negative on resection, or Oncotype DX was evaluated before completion of HER2 studies. Of note, the UC cohort had a nearly identical percentage of HER2-positive tumors (3.5%). The lesser importance of tumor size in the model may reflect its absence from the Oncotype DX scoring algorithm, despite its established clinical significance for staging and prognosis. The contribution of age to the model highlights the different risks that are associated with breast cancer progression in younger or older patients. Age's role underscores the biologically aggressive nature of tumors in premenopausal women, which often influences treatment decisions. The histologic subtype, although not a top feature in the model, can alter therapeutic sensitivity and recurrence risk, emphasizing that certain subtypes, such as lobular carcinoma, may require different treatment considerations. ER status was not a significant feature, as this study focuses only on ER-positive tumors. Additionally, nodal status, a critical staging factor, was not available in the dataset and included in the model, despite its importance in clinical decision-making, particularly for patients with few positive lymph nodes who receive local radiation. In summary, the logistic regression model's feature importance aligns with the clinicopathologic factors oncologists often consider when formulating treatment plans and prognoses for breast cancer patients.

In further comparative analysis using AUC scores and the Mann-Whitney U test with Bonferroni correction, OncoDHNet, the logistic regression model, and the combined model were evaluated on an internal held-out test set and an external UC cohort. In this analysis, the internal test set revealed significant differences. The OncoDHNet model significantly outperformed the logistic regression model, and the combined model demonstrated a marked improvement over OncoDHNet. Furthermore, the combined model showed substantial superiority to the logistic regression model. In the external UC cohort, significant performance differences were observed as well: OncoDHNet was notably better than the logistic regression model, and the combined model was significantly more effective than OncoDHNet, albeit less substantial than within the internal cohort. The combined model also outperformed the logistic regression model with a significant margin. These findings highlight that the relative performance advantages observed in the internal dataset were, in fact, replicated and magnified in the external cohort. The consistent superiority of the combined model in both test sets suggests a robust, generalizable advantage. It is important to note that both datasets are relatively small, and the observed variability underscores the need for further validation on larger, more diverse datasets in future work.

In collaboration with expert pathologists, OncoDHNet was qualitatively evaluated using attention maps that visually highlight regions key to its predictive analysis (27, 32). The attention maps highlight specific tissue areas within WSIs that the model deems critical in classifying the malignancy's nature and severity of disease. These visualizations provide pathologists with an intuitive understanding of the model's focal points, thereby demystifying the AI's operational logic and providing an opportunity for further refinement and verification. For example, a preliminary review of these visualizations for some "false positive" and "false negative" cases identified a few potential confounding factors such as dense lymphocytic infiltrates, biopsy site changes, extensive ductal carcinoma in situ, or a small amount of invasive carcinoma. Such transparency in the AI decision-making allows pathologists to confirm the model has appropriately selected areas of invasive carcinoma and facilitate its intergradation into clinical workflows.

A thorough review of electronic medical records for patients classified as low breast cancer recurrence risk by Oncotype DX but high risk by our model revealed significant implications for breast cancer prognosis. Of the 36 patients in the internal test set identified as high risk by our model and low risk by Oncotype DX, two experienced breast cancer recurrence. This finding suggests that our AI model has higher sensitivity in detecting high-risk cases compared to traditional genomic assays like Oncotype DX, capturing subtle, clinically relevant patterns potentially missed by current tests. Such enhanced risk prediction is crucial, as it can directly influence patient management, leading to more personalized and potentially life-saving treatment strategies. For example, patients who might otherwise be categorized as low risk and potentially undertreated could receive more aggressive therapy aligned with their true recurrence risk as identified by the AI model, thereby potentially saving lives through timely and targeted intervention.

This study has laid important groundwork for applying machine learning tools to the prognostication of breast cancer by determining low and high breast cancer recurrence risk categories. However, there are limitations to this study, as the current methodology, focused predominantly on histopathological features from WSIs and limited clinicopathologic features, which may not yet fully capture the complexity needed to predict the breast cancer recurrence score. This categorization is crucial as it guides the use of chemotherapy and helps in risk stratification, yet the proposed model does not integrate the additional clinical, pathologic, and molecular markers such as Ki67 scores and Extensive Intraductal Component (EIC) that are predictive of breast cancer recurrence. Additionally, the study does not extend to prognostic predictions such as time and likelihood of disease recurrence, metastasis, survival rates, and response to specific treatments, which are vital for comprehensive treatment planning. It also does not consider other significant features, such as lymphovascular space invasion, positive margins, or lymph node status which are known to influence the risk of recurrence and prognosis. Furthermore, menopausal status can be integrated to refine the model further, given the impact of menopausal status on tumor biology and response to treatment. By including additional genetic biomarkers, immunohistochemistry results, and patient demographic factors, the model training could be refined to improve predictive accuracy. The final goal is to enhance the current model's prediction accuracy while maintaining its practical accessibility.

Moreover, with few exceptions, the inference is drawn mostly from a single histological slide per patient (984 slides from 975 patients), which may not represent the heterogeneity accessible in a multi-slide analysis, a common practice in clinical settings. On the other hand, the Oncotype DX test is also limited by sampling bias – with a single core needle biopsy often serving as the source for testing. Future studies on AI models incorporating 3-dimensional, multi-slide analysis are needed and may prove better equipped to capture intra- and inter-tumoral heterogeneity that cannot be gleaned from a single biopsy site.

Finally, we plan to test this multi-model approach in clinical practice as part of a decision-support system for pathologists and oncologists. A prospective clinical trial can assess the utility of the proposed multi-model in predicting breast cancer recurrence risk and on patient health outcomes, which could be useful for understanding the model's performance in a real-world healthcare setting, specifically in terms of its' potential to reduce treatment delays and cost of care. Our future work aims to evaluate the impact of AI models on the accuracy and efficiency of breast cancer recurrence predictions and to show their efficacy in reducing reliance on costly genetic testing and ultimately improving personalized patient management in breast cancer care for diagnosis and prognosis.